# Feature anomaly detection system (FADS) for intelligent manufacturing

Anthony Garland*, Kevin Potter, Matt Smith

Sandia National Laboratories

* corresponding author. agarlan@sandia.gov

## Abstract

Anomaly detection is important for industrial automation and part quality assurance, and while humans can easily detect anomalies in components given a few examples, designing a generic automated system that can perform at human or above human capabilities remains a challenge. In this work, we present a simple new anomaly detection algorithm called FADS (feature-based anomaly detection system) which leverages pretrained convolutional neural networks (CNN) to generate a statistical model of nominal inputs by observing the activation of the convolutional filters. During inference the system compares the convolutional filter activation of the new input to the statistical model and flags activations that are outside the expected range of values and therefore likely an anomaly. By using a pretrained network, FADS demonstrates excellent performance similar to or better than other machine learning approaches to anomaly detection while at the same time FADS requires no tuning of the CNN weights. We demonstrate FADS' ability by detecting process parameter changes on a custom dataset of additively manufactured lattices. The FADS localization algorithm shows that textural differences that are visible on the surface can be used to detect process parameter changes. In addition, we test FADS on benchmark datasets, such as the MVTec Anomaly Detection dataset, and report good results.

# Background

## Introduction

Anomaly detection is important for many types of industrial automation and part quality assurance, and while humans can detect anomalies in components given a few examples, designing a generic automated system that can perform at human or above human capabilities remains a challenge (Ruff et al., 2021). For this work, anomaly detection is focused on unsupervised part defect detection based on optical images. Ideally such a system should be generic and not require hand-crafted features or region identification algorithms which often are costly to design (since an engineer must spend a significant amount of time developing the algorithm). Additionally, hand-crafted algorithms are often brittle and do not perform well when minor changes to the input images are applied. Anomaly detection is often called out-of-distribution detection since the goal is to identify when the inputs vary significantly from the expected nominal distribution (Ruff et al., 2021; Yang, Zhou, Li, & Liu, 2021). For high dimensional data, such as images, determining the nominal distribution is not trivial because some (often significant) variation in the nominal images is expected and separating the natural variation from meaningful changes in the input is difficult. Since images of manufactured components are complex and contain a significant amount of information, the process of how to map an image to a 'expected distribution' is not obvious.

## Anomaly Detection for Additive Manufacturing

Anomaly or defect detection is commonly used in industrial manufacturing (Barari, Tsuzuki, Cohen, & Macchi, 2021; Hao, Lu, Cheng, Li, & Huang, 2021; G. Kim, Choi, Ku, Cho, & Lim, 2021). However, anomaly detection can be especially difficult for low volume or one-off volume production which is often the case when using additive manufacturing (AM) (Thompson et al., 2016). 3D printers are essentially complex electro-mechanical-cyber robots that can have any number of physical, electrical, or software problems which makes the exposed risk hyperspace large when using 3D printers to produce parts that will be used in production (Herzog, Seyda, Wycisk, & Emmelmann, 2016; Lehmann et al., 2021; Paolini, Kollmannsberger, & Rank, 2019; Sanaei & Fatemi, 2021). While *in-situ* monitoring and digital twinning of the 3D printer can help mitigate risks(Everton, Hirsch, Stravroulakis, Leach, & Clare, 2016; Mi et al., 2021; Qi, Chen, Li, Cheng, & Li, 2019; Scime & Beuth, 2019), these techniques cannot detect all anomalies and often times physical access to the printer maybe limited (e.g. when using 3D printing as a service) and the exact software settings used to command the printer are closely held trade secrets that vendors are not willing to share. This lack of information sharing makes *in-situ* monitoring and guarantees about the software used or physical state of the printer unavailable for quality assurance, and instead the end-user can only inspect the final product (Seifi, Salem, Beuth, Harrysson, & Lewandowski, 2016).

Despite having a large risk space, AM has many benefits. Some key benefits of AM are first, the ability to produce low volume batches of parts where setting up an entire manufacturing line would be extremely expensive or producing a few parts in a machine shop would require hours of work by skilled. Second, AM enables producing parts that are impossible to make via traditional subtractive processes and often significantly reduces the amount of waste material. During product development, parts are often additively manufactured as functional prototypes, and if the prototype is successful, then more of the part is needed for the initial production runs of the system. As a result, only a few examples of correctly fabricated parts may exist. Low volume anomaly detection can be a helpful tool in this case. Anomaly detection for AM parts is important first to ensure that the part's geometry is correct and second to ensure that the process parameters of the 3D printer have not drifted from the calibrated state. While certain aspects of a AM components cannot be inspected from the outside and can only be inspect with computer tomography (such as internal hidden features like cooling channels) (du Plessis, Yadroitsava, & Yadroitsev, 2020), optical inspection of the outer appearance of the part is a common first step in a chain of qualification tasks for AM parts. Additionally, while obvious flaws that are visible from the outside need to be detected, other subtle changes in the part quality are much more difficult to detect. For example, as a 3D printer ages, the output energy of the laser may degrade slowly which will impact the quality of the sintering process for a laser powder bed fusion process (a similar problem is discussed in (Y. Kim, Lee, & Kim, 2021)). In an adversarial scenario involving high consequence applications and parts, a malicious actor could attempt to secretly modify the firmware of the 3D printer to purposely limit the laser power below the commanded laser power setting. In both scenarios, a "trust but verify" approach is needed to confirm that the parts were made correctly and are of the same quality as the parts printed during the design prototyping stage (Hensley et al., 2021). While techniques exist for monitoring the performances of printers, such as printing and mechanically testing witness coupons in every build (Salzbrenner et al., 2017), being able to simply take an image of a part and determine if the build was in the nominal regime would be valuable (Li, Jia, Yang, & Lee, 2020).

## Anomaly Detection Methods

An important part of human anomaly detection is the realization that humans fit anomaly detection into a broader picture of object identification and expectations of the structure of the world. Unlike some anomaly detection systems which are only given examples of the nominal cases images and nothing else, humans are given a lifetime of data about all sorts of objects and then a few nominal images as a very small subset of their broader experiences. As a result, humans can approach anomaly detection with a rich set of feature identifiers that map to something other than the class of the nominal images. Anomalies are identified as an object or section of an object which maps to something that the person has seen before which is not the nominal object. Within anomaly detection, the core problem is both mapping high dimensional nominal images to some small region in a reduced order space and ensuring that the mapping does not degenerate and result in all (nominal and anomalous) images mapping to the same region in the reduced order space (Ruff et al., 2021; Ruff et al., 2018). The important observation is that having a very broad set of experiences and predetermined features is likely very helpful for anomaly detection.

An additional challenge for machine learning anomaly detection is the ability to detect not only semantically different objects (yellow fire engine vs. yellow school bus) but also textural differences within the object (a shiny red fire engine vs a pastel red fire engine) (Djavadifar, Graham-Knight, Körber, Lasserre, & Najjaran, 2021). Ideally a machine learning algorithm could learn how to identify both types of anomalies without the need for hand crafting features. In many cases it is not clear what type of anomaly to look for *a prior (Smith, Laursen, Bartanus, Carroll, & Pataky, 2021).* For example, when investigating a 3D printed lattice with a repeating structure, should an anomaly detection system focus on the repeating pattern of the lattice (potentially a type of texture at a higher level and likely an indicator of process parameter change based on our previous work (Garland et al., 2020)) or the surface quality which influences the part quality as well (based on another previous work (Roach et al., 2020))?

Recently, deep learning has been used for anomaly detection in a wide variety of applications, and deep learning shows impressive results on anomaly detection tasks. A class of deep learning anomaly detection methods train auto-encoders or generative adversarial networks to learn the distribution of nominal images by learning a low dimensional representation of the image and then regenerating the input image from the low dimensional representation (Donahue et al., 2019; Haselmann, Gruber, & Tabatabai, 2018; Sabuhi, Zhou, Bezemer, & Musilek, 2021; Schlegl, Seeböck, Waldstein, Langs, & Schmidt-Erfurth, 2019; Schlegl, Seeböck, Waldstein, Schmidt-Erfurth, & Langs, 2017; Tuluptceva, Bakker, Fedulova, Schulz, & Dylov, 2020). At inference time, the generative system attempts to reproduce the input image. If the system's output is a poor representation of the input image (because the nominal images had no data in this regime) then the difference of the input image compared to the output image is high and the image is labeled an anomaly (Donahue et al., 2019). Another set of anomaly detection techniques seeks to learn meaningful features for anomaly detection using contrastive loss by applying geometric transforms to the nominal images (Golan & El-Yaniv, 2018). Although this technique is helpful, for many application domains, for other domains (like additively manufactured lattices) it is not clear how to generate meaningful geometric transforms.

In application areas, outside of anomaly detection, machine learning solutions rely extensively on transfer learning (Zhuang et al., 2021). Transfer learning involves training a neural network on an axillary task which often has large datasets available, and then transferring the knowledge to a new neural network performing a similar but slightly different specialized task (Tan et al., 2018).

Deep convolutional neural networks (CNN) are often used for image classification where the network learns high level features (such as textures) in the first few convolutional layers and sequentially deeper layers learn more and more abstract representations of the input image. The final convolutional layer is aggregated into an N dimensional vector where each of the N dimensions corresponds to the probability of an object type being in the image.

In the deep one class classification methods for anomaly detection, CNNs are trained to map all the nominal images to a small region with center, C, in a hyperspace (Ruff et al., 2018; Song, Liu, Huang, Wang, & Tan, 2013). At inference time, the input image is passed through the network and the output vector is computed and compared to the center, C, by calculating some distance metric. The distance of the output vector from the center, C, is interpreted as a measure of the likelihood that the image is an anomaly. However, the core challenge with the deep one class methods is that a mapping to a small hyperspace degenerates into mapping everything (both nominal and anomalous images) to the same location on the hyperspace. Some sort of regularization is needed to prevent this collapse. A variety of techniques have been proposed to solve this problem with varying degrees of success (Goyal, Raghunathan, Jain, Simhadri, & Jain, 2020; Ruff et al., 2021; Ruff et al., 2019).

Based on the previous discussions, a potential solution for some of the challenges in anomaly detection is to leverage pretrained networks, but how to reuse a pretrained network for anomaly detection is not clear. Our proposed algorithm (Feature anomaly detection system) offers one potential solution.

## FADS (Feature anomaly detection system)

In the deep-one-class classification anomaly detection paradigm, an unsupervised deep learning anomaly detection system should map the nominal high dimension inputs (images in this case) to some small region or regions in a hyperspace, but anomalies will map to some other region in the hyper space.

This hyperspace needs to encode not only textural information but also semantic information about the nominal objects. For example, if the nominal data is cats, then a dog is an anomaly, but a cat with grass as fur is also an anomaly. Therefore, the latent representation used for the anomaly detection system needs to consider both the textural and semantic differences in objects. In the context of a convolutional neural network (CNN) this distinction translates into examining the outputs of the early features which can capture textural information about an input image and examining the deeper layers which capture the semantic information. A general critique of deep one-class anomaly detection algorithms is that the 'deep' representations are unable to detect small textural differences (Ruff et al., 2018). Additionally, anomaly detection systems should detect defects that are only a small part of a nominal component.

Based on these observations, we propose Feature Anomaly Detection System (FADS) which uses a pretrained CNN to learn the statistical distribution of the convolutional filter activations by passing the training data through the CNN and recording the activations. This distribution is then the basis for classifying a new image as nominal or an anomaly. FADS is similar to deep one-class classification like algorithms in that all the nominal images are mapped to a single location in a hyper-space, but FADS differs in that all convolutional features in all the layers are used during this mapping and that the model is pretrained. By using all the convolutions at all levels of feature detection, we expect FADs to capture not only textural anomalies based on features in the first few layers but also semantic anomalies in the final convolutional layers.

In formal mathematical description, we are given a pretrained CNN model $f$ which has $I$ convolutional filters and given an image, x, pass the image through the model,

$$\bar{y} = f(x), \qquad (1)$$

and record the convolutional filter activations $\phi_{i,k,l}$ where $i$ is the *ith* convolutional filter and *k* and *l* are the dimensions of the 2D activation map. Discard $\bar{y}$ which is the output vector normally used for image classification. Since each convolutional filter produced a 2D map of activations, an aggregation function $f_a$ is used to summarize the activation map to a single number,

$$\phi_i = f_a(\phi_{i,k,l}) \qquad (2)$$

where $\phi_i$ is the aggregated *ith* convolutional filter activation of the image, *x*. We investigated min, max, and mean as potential aggregation functions. The nominal images $X_N$ are passed through the network and the activation maps are summarized using equation (2) which results in $\phi_{i,x_j}$ where $x_j$ indicates the *jth* image in the nominal set. The average and standard deviation of the aggregated *ith* feature is found across the *n* images in the training set.

$$\phi_{i,\bar{x}} = \frac{\sum_j^n \phi_{i,x_j}}{n} \qquad (3)$$

$$\phi_{i,\sigma} = \sqrt{\frac{\sum_j^n \left(\phi_{i,x_j} - \phi_{i,\bar{x}}\right)^2}{n}} \qquad (4)$$

The mean activations, equation (3), represents the central location in hyperspace for the nominal images, and the standard deviation, equation (4), represents the expected nominal variation of each dimension in the hyperspace. Both the mean and standard deviation vectors are visualized in Figure 1c.

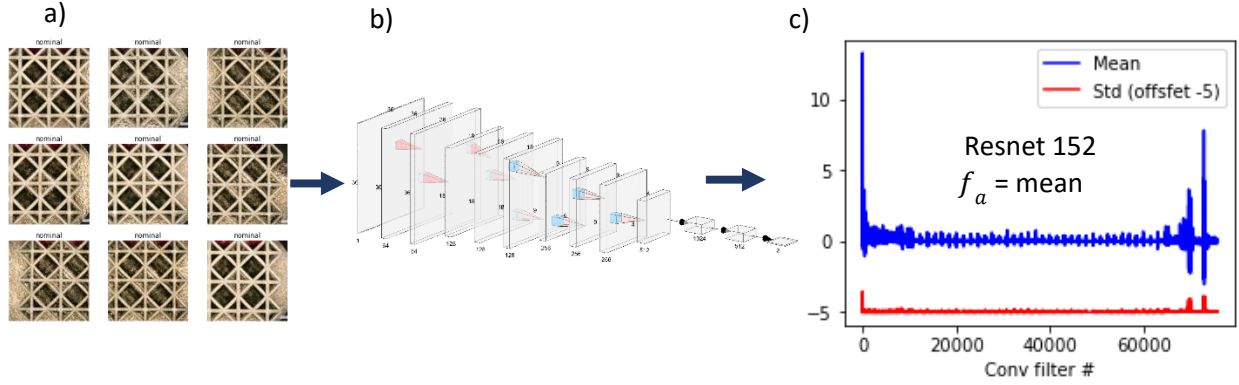

Figure 1 a) Example training data b) The image data is passed through a CNN where the activation maps are aggregated into a single number using $f_a$ for each training image. c) The mean and standard deviation of the output of the aggregation function across the input images plotted against convolutional filter. This plot represents the output from equations (3) and (4).

For inference, we are given a test image $\tilde{x}$ which may come from the anomaly set $\mathbb{A}$ or from the nominal set $\mathbb{N}$. Like before, the image is passed through the network and the features maps are aggregated using Equations (1) and (2) which gives $\phi_{i,\tilde{x}}$. Next, we calculate a difference vector $r$, which is the number of standard deviations from the average activation found during training for each convolutional filter,

$$r_i = \frac{|\phi_{i,\tilde{x}} - \phi_{i,\bar{x}}|}{\phi_{i,\sigma}}. \qquad (5)$$

The r-vector is an indicator of how much each filter activation in the test set deviates from the filter activations in the training set. High deviation from the training set would indicate that something in the image is different in this image from the training set. Therefore, the anomaly score is either finding the max deviation of the value r-vector $s_{max}$, or the value of the 90[th] percentile of the r-vector, $s_p$, and is shown in Equations (6) and (7).

$$s_{max} = max(r) \qquad (6)$$
$$s_p = percentile(r, 90) \qquad (7)$$

Calculating the p-2 norm of the r-vector would result in the Mahalanobis distance metric which is commonly used in anomaly detection; however, empirically we found that this distance metric performed poorly and that Equations (6) and (7) performed better. Figure 2 shows a flow chart of the inference pipeline. Figure 3 shows a visualization of the magnitude of the r-vector, Equation (5), on the test set. For most images in the test-set, the magnitude of the r-vector is much larger at one location or many locations for the anomalous data, in subfigure b), compared to the nominal data, in subfigure a), and therefore provides a visual aid in understanding why Equations (6) and (7) work.

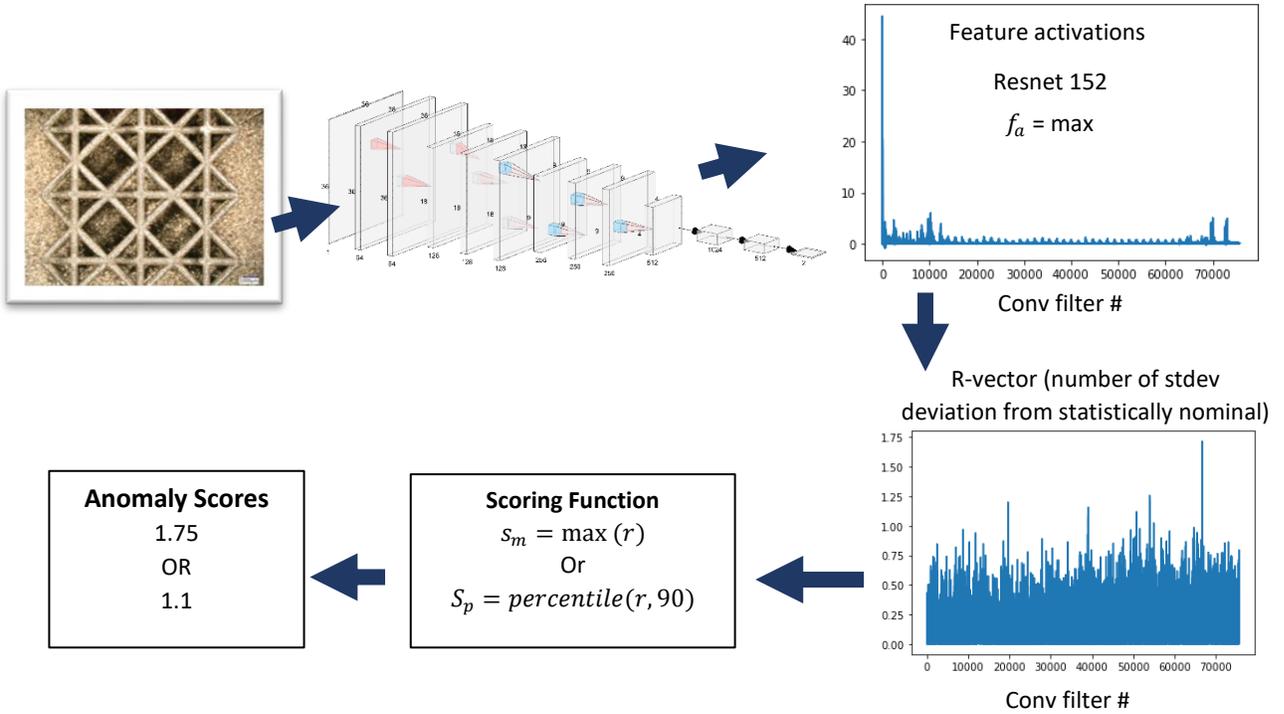

*Figure 2 shows the inference flow pipeline. The input image is passed through the CNN, the activation maps of each conv filter are summarized using $f_a$, the r-vector is calculated (Equation (5)), the scoring function converts the R-vector to a single anomaly detection score.*

Like all machine learning algorithms, FADS relies on setting hyperparameters to achieve the best results. For the case of unsupervised anomaly detection, tuning hyperparameters is difficult because there is no way to measure the effectiveness of the algorithm for a particular set of hyper-parameters since no examples of anomalies are known. In practice, often a few examples of anomalies are known and can be used to help set hyper-parameters. FADS is sensitive to the selection of the pretrained network and input image size used in Equation (1) since these both effect which convolutional features will be activated. Additionally, the aggregation function, Equation (2), and the selection of the scoring function, Equation (6) or (7), can influence the results. Empirically, we found the *max* function works well for Equation (2) and the scoring function (Equation (6)) for most applications. To make FADS insensitive to network and image size selection, FADS is run several times with different image sizes and pretrained CNNs, the outputs are normalized and then combined to give a single anomaly score.

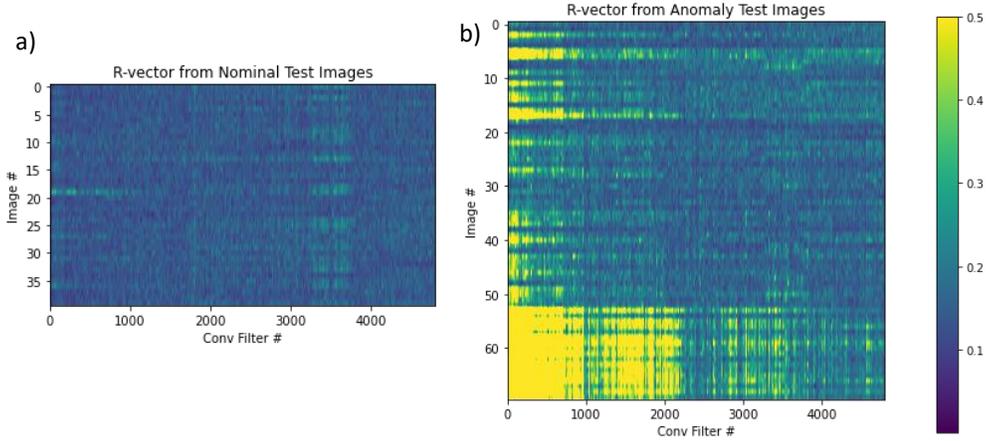

Figure 3 a) and b) show the r-vector for each image in the test set from running FADS on the hazelnut images in the MVTec dataset. a) shows the r-vectors for the nominal images and a lower magnitude of values for most images while b) shows the r-vector for the anomalous images with a higher magnitude of values.

We normalize each run of the FADS network by the average anomaly score $s_{max}$ or $s_p$ of the training data on itself. Given Z FADS models with different hyperparameters, calculate the average anomaly score $\bar{s}_{a,z}$ for all the images in the training set. $\bar{s}_{a,z}$ is the average score (indicated by the bar) of $z$th FADS model that uses scoring method $a$ (which is either the max or percentile method) when the training images are scored on their own FADS model.

At inference time when given input image, $\tilde{x}$, calculate the anomaly score from each FADS model $\tilde{s}_{a,z}$, normalize by dividing by $\bar{s}_{a,z}$ and then average the resulting scores together,

$$s_{a,final} = \frac{1}{Z}\sum_{z=1}^{Z}\frac{\tilde{s}_{a,z}}{\bar{s}_{a,z}} \qquad (8)$$

where $s_{a,\text{final}}$ is the final anomaly score using scoring method $a$. This method is essentially creating an ensemble of the results from each model. It also has a self-normalization property which is helpful in real world examples with zero examples of anomalies. Whereas academic datasets, like MVTec (Bergmann, Batzner, Fauser, Sattlegger, & Steger, 2021; Bergmann, Fauser, Sattlegger, & Steger, 2019), enable calculating a performance metric for a variety of different hyperparameters, real world applications often have no examples of anomalies and therefore the optimal threshold between anomaly and nominal needs to be inferred from the algorithm itself. The normalization from Equation (8) results in a natural decision boundary between anomalies and nominal scores at a value of one.

### Visualization

The regions that are causing the high anomaly score can be visualized by calculating how to perturb the input image in order to lower the anomaly detection score. Although many methods for explainability of the output prediction could be used, guided back propagation (Samek, Montavon, Lapuschkin, Anders, & Müller, 2021) worked well. In essence, the visualization tries to identify how to make the input image more 'normal'. Given an image $\tilde{x}$, compute the r-vector, Equation (5), and then calculate the loss of the r-vector with respect to the zero vector, **0**, using some loss function $l$. For this work, mean square error was used as the loss function, $l$. If the r-vector was the zero vector, then the image would represent an ideal anomaly-free image.

$$loss_{\tilde{x}} = l(r_{\tilde{x}}, \mathbf{0}) \qquad (9)$$

Once a loss value is computed, a standard visualization method can be used. For this work, we used guided-back-propagation to visualize the regions of the input image which are causing an high anomaly score (Lee, Jeon, & Lee, 2021; Springenberg, Dosovitskiy, Brox, & Riedmiller, 2014). In practice, using equation (9) does not result in good visualizations. Instead, we found that minimizing the loss of the top 10 percent of dimensions in the r-vector works well. These highest 10 percent of values in the r-vector represent the dimensions that deviate the most from the expected distribution of activations and better aligns with the anomaly scoring methods in Equations (6) and (7).

## Experimental Setup

### MVTec

To evaluate the performance of FADS, we first applied FADS on the MVTec dataset which is a dataset of 15 categories of images from a manufacturing environment (Bergmann et al., 2019). MVTec has the advantage of having labels, including somewhat rough pixel-wise labels of where anomalies are located.

Our localization method can highlight pixels that are causing a high anomaly score, however the MVTec localization maps consist of general regions that have anomalies rather than individual pixels. As a result, we used an average pooling function to calculate the average number of pixels above a threshold, $\phi^t$, in a region of the image. The region size is *n* by *n* based on the average pooling functions parameters. If the output of the average pooling is above another threshold, $\rho^t$, then the region is labeled as an anomaly.

### Lattices

The second dataset consisted of additively manufactured (AM) lattices where for some of the lattices the laser power and scan speed were purposefully varied to change the amount and quality of material deposited when creating the lattice (Garlea et al., 2019). In addition, some of the lattices with nominal print settings had one or two struts missing per unit cell to simulate a broken or missing strut. Three build plates were used where the build plates one (P-1) and three (P-3) had 25 nominal lattices. The second plate (P-2) had 5 nominal lattices and 18 lattices with anomalies. The training and test data split was setup to ensure that some of the nominal lattices from plate two were in both the training and test set. Since the dataset was small, we performed a stratified k-fold analysis with 7 folds where the stratification class was the build plate.

In addition, from our previous works (Garland et al., 2020; Roach et al., 2020), we suspected that surface roughness would be a key indicator of the process parameters used during the build process. As a result, we hypothesized that different image types and different lighting conditions would reveal different surface features causing different shadows. For each lattice we took 5 images from the top down with different lighting conditions (Cheng & Tsai, 2021) using a Keyence Microscope. Additionally, some features (e.g., broken/missing struts) are not visible from the top view. Therefore, we took four additional images from the four orthogonal side views using a 4.3 megapixel FLIR grasshopper3 digital camera. In both cases A fixture was used to align the lattice samples within the imaging set up to maintain as much consistency as possible. All images were converted to grayscale before processing with FADS. Figure 4 shows the process parameter map and the different image views in color before the images were converted to grayscale.

Since most of the anomalies in the lattices are changes in textures, we set 'mean' as the aggregation function, $f_a$, of each convolutional activation map. We found the 'max' aggregation function is generally good at detecting physical feature changes, and the 'mean' function is better at detecting subtle textural differences. Whereas large but highly localized changes (e.g. missing struts) in an image would cause a single large activation of a convolutional filter at one location that is readily detected by the max aggregations function, textural differences (e.g. change in process parameters that affects the entire lattice) will not show a large max aggregation, but will show even subtle differences in their mean aggregation

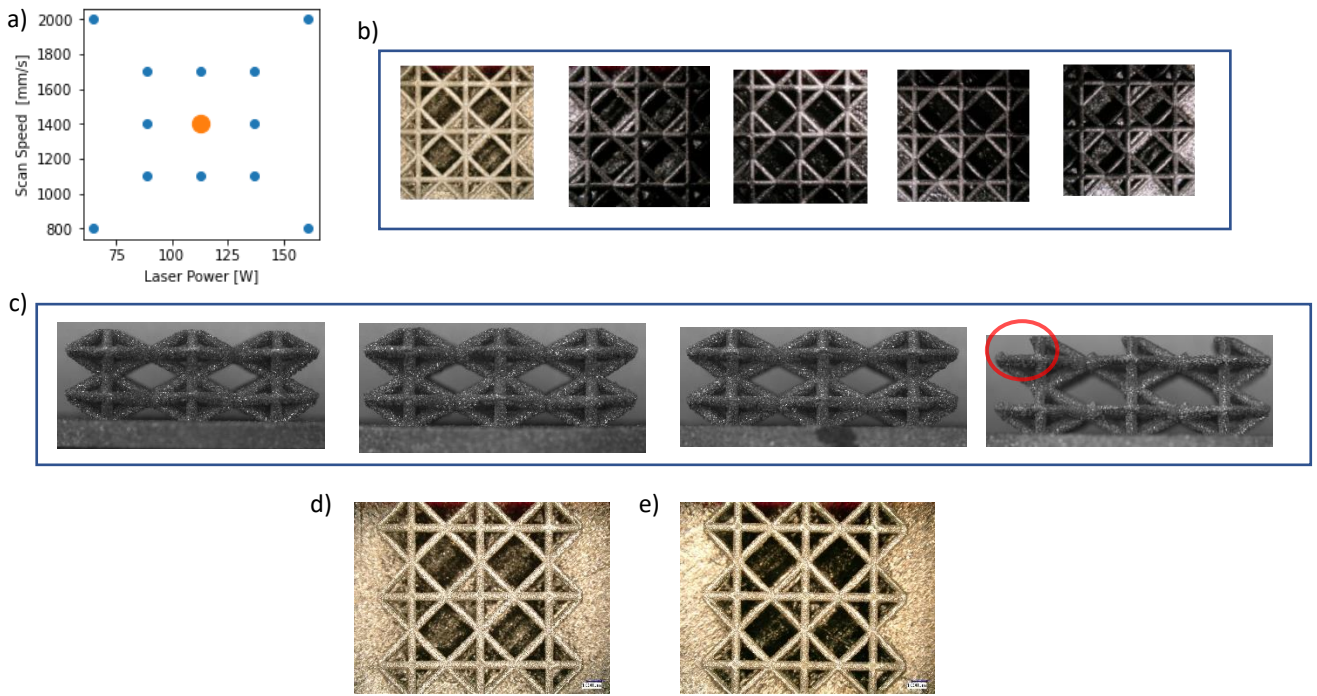

Figure 4. a) The process parameters used to print the lattices. The orange circle represents the nominal print settings while the surrounding blue circles represent the process parameters used to generate anomalous lattices. b) The top down images taken with different lighting conditions to reveal different surface features. c) The four orthogonal side views. The image on the right shows a missing strut at the circle. d) shows a nominal lattice while e) shows lattice with incorrect process settings. All images were converted to grayscale before processing with FADS.

## Results

The MVTec results are shown in Figure 5. Figure 5a shows the 15 different images types in the MVTec dataset. The receiver operator curve (ROC) shows the tradeoff between true positive and false positive values for different anomaly threshold values. The area under the receiver operating curve (AUC ROC)

(Hand & Till, 2001) values summarizes the ROC curve to a single value where a value of one for the AUC ROC would represent perfect distinction between nominal and anomalous images while an AUC ROC of 0.5 would represent guessing which images are nominal and anomalous. The FADS algorithm gives an average AUC ROC of 0.93 with several categories having perfect results. Figure 6 shows the output of the FADS localization algorithm.

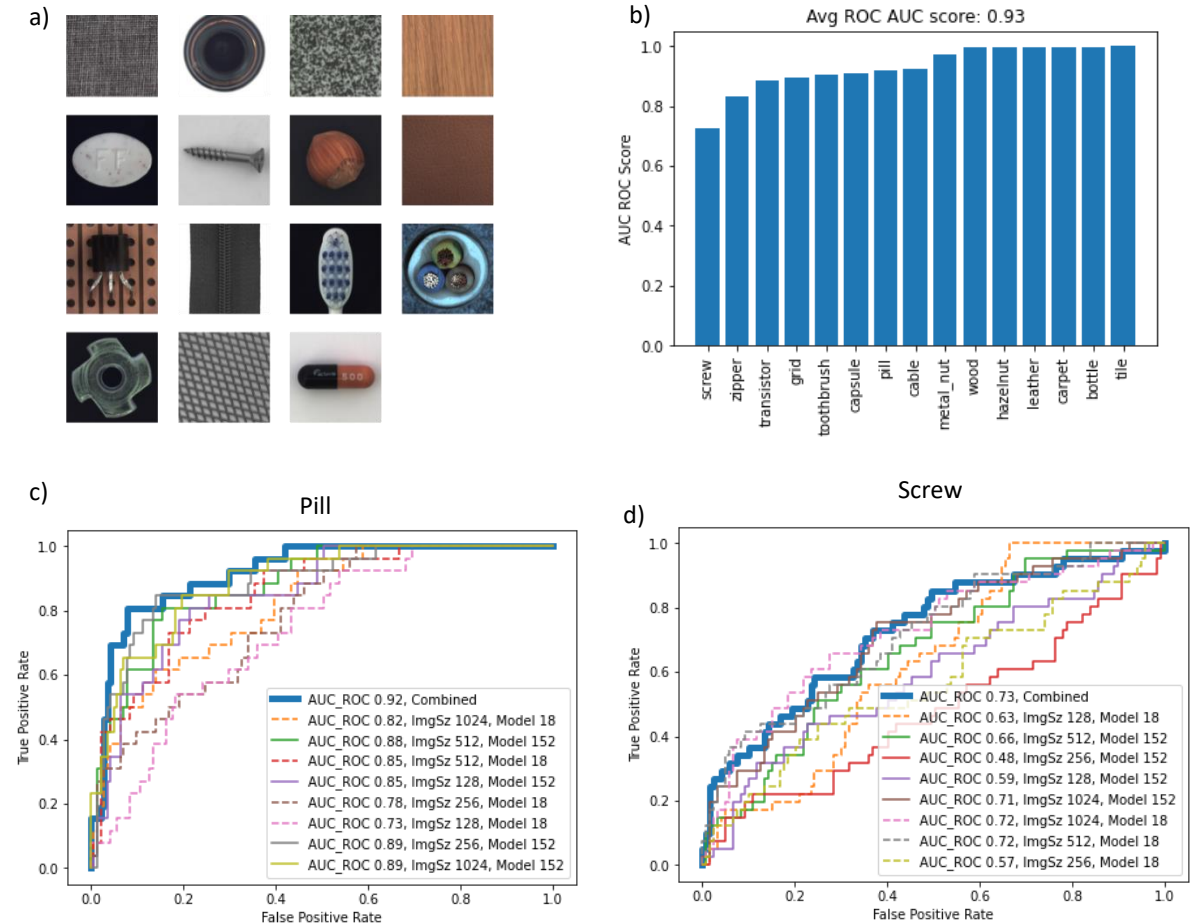

Figure 5 a) Nominal images from the 15 images types in the MVTec dataset. b) shows the AUC ROC for each class in the MVTec dataset. c) and d) show the ROC for a variety of the hyperparameters for the Pill and Screw image types. While the deepest CNN with the largest image size performs well, an improvement is still observed by ensembling the output results.

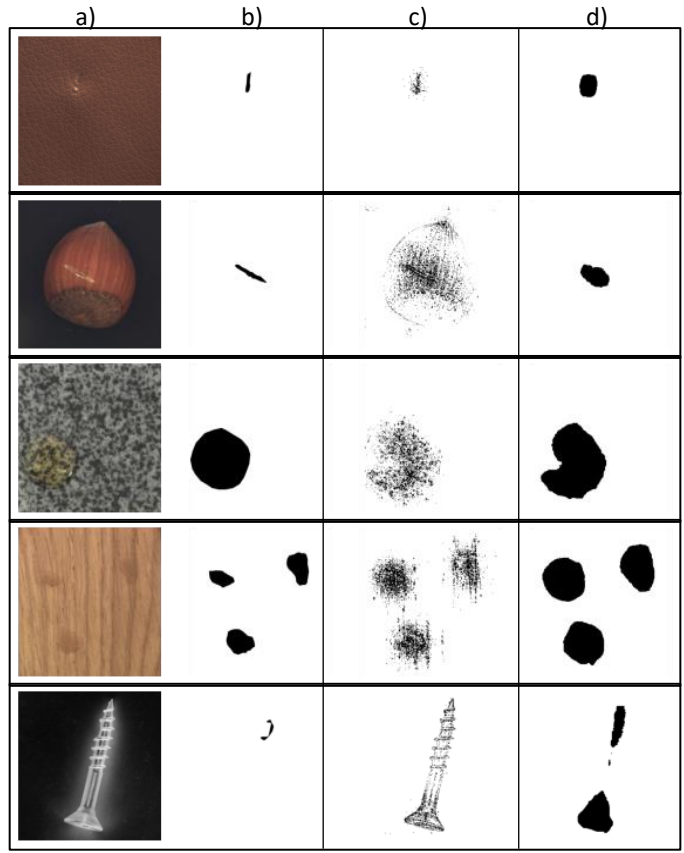

*Figure 6 a) Shows the input image, b) shows the ground truth segmentation maps c) shows the raw output of FADS localization algorithm d) shows the FADS localization after passing through a pooling layer. While the first four rows show good localization, the last row with the screw shows a failure case.*

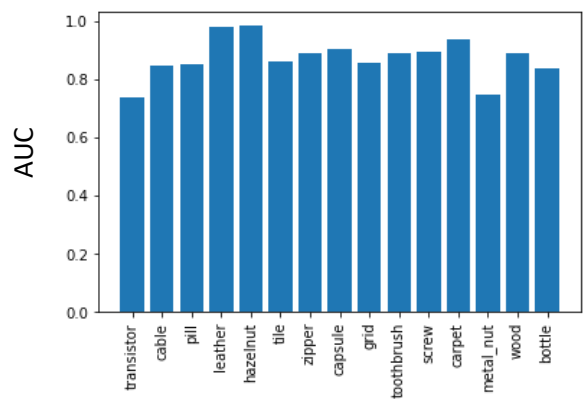

*Figure 7 Pixel ROC AUC for MVTec .*

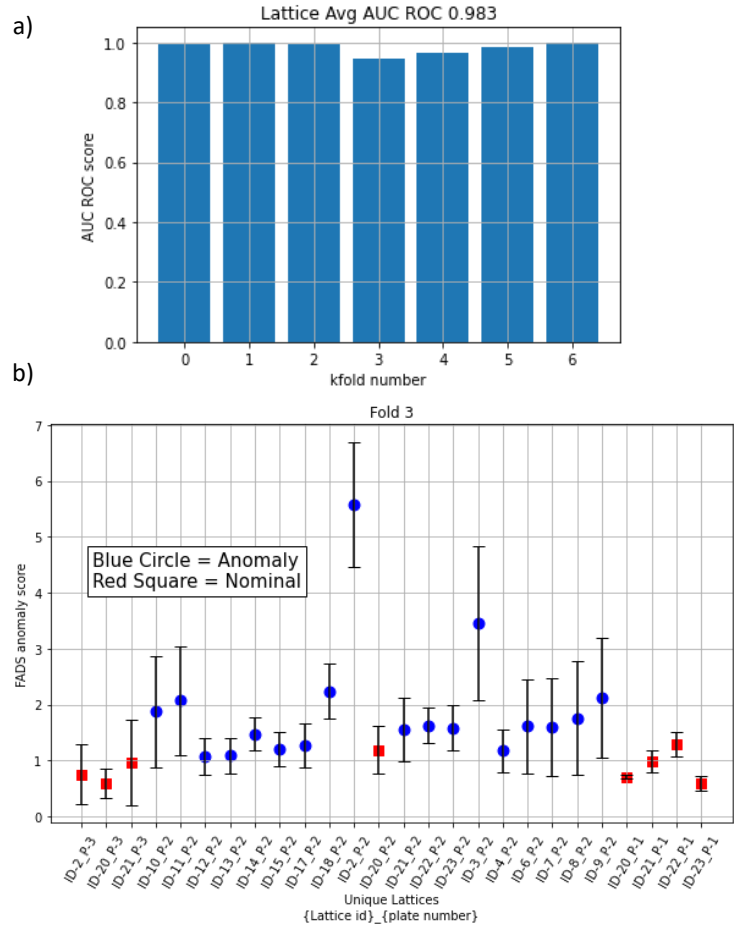

Figure 8 a) AUC ROC scores for each fold. b) shows the average FADS score from all the images of a particular lattice for fold 0 as computed using Equation (8) where the hyperparameter change is different image views. The error bars indicate one standard deviation from the average anomaly score from the different image views.

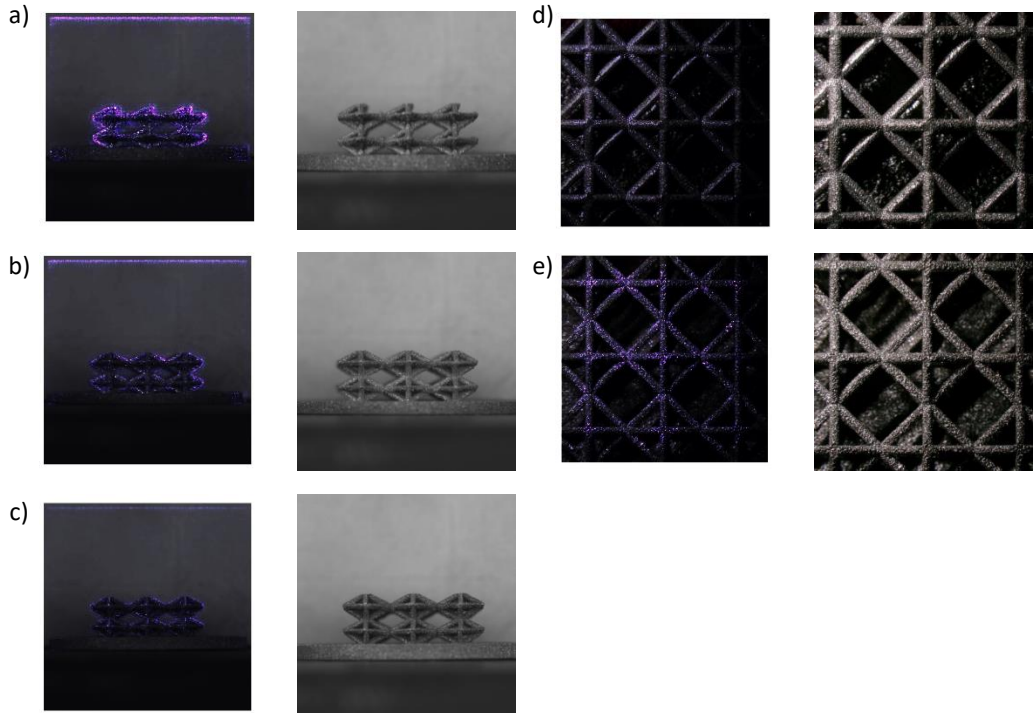

*Figure 9. Visualization of the regions causing high anomaly scores. On the left of each image pair is the localization overlayed the original image. On the right of each image pair is the original image. Images a), b), and e) are anomalies while c) and d) are nominal and therefore the localization highlights are sparse and of low intensity.*

## Discussion

The results show that the FADS algorithm is near state of the art for generic anomaly detection but has the added advantage of requiring no training of the NN's weights. In addition, the localization of defects performs well and can provide helpful information about the exact location of a defect in many instances. FADS performed well on both the MVTec and the custom lattice dataset. The localization visualizations from the lattices in Figure 9 show that FADS is using the actual lattice in the image for anomaly detection and not some spurious correlation in the background. Additionally, Figure 9 subfigures b)-e), seem to confirm our hypothesis that textural differences are the main difference between the nominal and anomalous lattices since subfigures b) and e) highlight specks all along the lattice rather than a specific location. We would also note that the difference in Figure 9 between b) and c) and also the differences between d) and e) are very subtle and that an untrained person would not be able to detect the anomaly which shows the strength of the FADS algorithm. Figure 9 subfigure a) shows how FADS highlights the outer shape of the lattice profile which is partially expected since the missing strut causes a significant change in the profile shape of the lattice. Ideally, FADs would highlight the exact location of the missing strut rather than the profile. We hypothesized before the experiment that changes in laser speed and power would result in not only textural changes that are visible on the surface of the lattice but also would result in small changes in strut thickness. We expect both to be used by the algorithm, yet the visualizations seem to indicate that textural changes on the surface of the lattice drive the anomaly detection score for most lattices unless there is a gross defect, such as a missing strut (Simoni, Huxol, & Villmer, 2021). The lattice dataset anomaly detection results seem to indicate that merely taking images of test artifacts built alongside the main components within the AM machine can be used as a simple quality control measure when coupled with the FADS algorithm.

The ensembling method in Equation (8) does improve the overall performance of the FADS algorithm and is shown in Figure 5 c) and d). While ensembling might not work well when rapid millisecond inference is required, since they require additional computational costs and thus time, in many manufacturing settings where inference is at a slower pace, the ensemble method works well. We would also note that inference speed for a single FADS model is similar to the inference speed for classification with the pretrained model. The additional overhead of computing the r-vector (Equation (5)) is minimal. In general, the deeper pretrained networks (e.g. ResNet152) performed better than shallower networks (e.g. ResNet18) which is expected since deeper networks have more convolutional activations to define the nominal distribution. Therefore, if ensembling cannot be used, then using a single deeper pretrained network is recommended.

The individual scores from the ensembling method can provide some measure of uncertainty, yet this measure should be used with caution based on the results in Figure 8b). In Figure 8b) the one standard deviation error bars extend passed the boundary between anomaly and nominal scores, which is approximately at 1, for both the nominal and anomaly lattices. For the lattice, the images are actually from different views which can view different regions of the lattice and therefore the underlying generating function is different. Because different generating functions are used to create the images, it could be expected that the spread of anomaly scores is larger. Similar to other anomaly detection methods, the anomaly score from a single FADS model, which is often some measure of the distance from the nominal position, does not provide meaningful measure of uncertainty other than the magnitude of the actual distance from the nominal position. Based on our observations of the FADS scores, we found that anomalies rarely had a score lower than 1. A more common error was for nominal images to have an anomaly score between 1 and 1.5. Equation (8) causes the expected threshold between anomaly and nominal to be at a value of one.

In contrast to deep one class classifications methods for anomaly detection, FADS may increase the dimensionality of the input image's representation when used with a very deep CNN with many convolutional filters. For example, an image with dimensions 256 by 256 (65,536 pixels) when passed through ResNET152 will result in greater than 70,000 aggregated feature activations. This indicates that the pretrained CNN acts as a transformation function of the input image to an alternative features space where it is easier to identify anomalies using simple functions like Equations (6) and (7).

FADS as several limitations. First, the approach assumes that the nominal data can be clustered in the space of filter activations and is bounded in some sense. While this might be the case for many application areas, having enough data to generate the true bounding can be difficult. For example, simple rotation of the screw could be a more significant change than an actual defect in the screw. The distance in the activation feature space could vary more between individuals that are both nominal than between a pair of nominal and anomalous images. A potential solution is to cluster the nominal images first and then to find the nominal activations for each group (Ghafoori & Leckie, 2020). At inference, an image is compared only to its nearest cluster.

A second limitation is the inability to guarantee that an image will be classified as nominal or an anomaly which limits the helpfulness of a few labeled examples (Ruff et al., 2019). For example, if FADS learns the nominal distribution from 100 images, then a new nominal image is presented, but falls out of the nominal distribution, then FADS has no simple way to correct this situation. One approach is to recompute the nominal distribution using the new image, but this doesn't guarantee that the new image will now be classified in the nominal range. A pre-clustering algorithm might also help with this limitation.

A general limitation of the lattice dataset is that all the anomalies were on a single build plate, and while nominal 4 or 5 lattices from build plate 2 (P-2 in Figure 8b)) were used in the 'training' data for each fold and 5 of the 7 folds had at least one nominal lattice from P-2 in the 'test' set, this might not be enough to conclusively prove that FADS is not generally identifying some difference between plates and identifying as anomalies all lattice that are not similar to plates P-1 and P-3. Although, we would note that several of the folds with perfect scores in Figure 8a) require FADS to correctly identify the one nominal lattice from P-2 in the test set.

## Summary and Conclusion

We present feature-based anomaly detection system (FADS) which uses the distribution of feature activations caused by nominal data when passed through a pretrained convolutional neural network to generate a statistical representation of the nominal data. This statistical representation along with the feature activations caused by a new input image can be used to classify the new image as nominal or anomalous. A loss function which computes the distance between an image's feature activations and the statistically nominal activations enables using guided back propagation to visualize the regions of the image which would need to change in order to lower the anomaly score. The FADS algorithm requires no optimization of neural network weights which may be attractive for some specialized applications where back propagation training of a NN is prohibited.

Our results on the MVTec dataset (average AUC ROC of 0.93) are near state of the art for unsupervised anomaly detection in images. In addition, we show the usefulness of the FADS algorithm (average AUC of 0.983 across 7 folds) on a new dataset comprised of additively manufactured lattices. The FADS anomaly localization algorithm showed that changes in laser power and speed result in in subtle changes in texture on the surfaces of the lattices which can then be used to identify anomalies.

## Acknowledgements

Sandia National Laboratories is a multimission laboratory managed and operated by National Technology & Engineering Solutions of Sandia, LLC, a wholly owned subsidiary of Honeywell International Inc., for the U.S. Department of Energy's National Nuclear Security Administration under contract DE-NA0003525. Laboratory facilities were provided in part by the Center for Integrated Nanotechnologies (CINT). This paper describes objective technical results and analysis. Any subjective views or opinions that might be expressed in the paper do not necessarily represent the views of the U.S. Department of Energy or the United States Government.

In addition, we would like to thank Ben White for imaging the lattices and Scott Jenson, Michael Heiden, and Jonathan Pegues for printing the lattices with different process parameters.